\documentclass[11pt]{article}

\usepackage[preprint]{acl}
\usepackage{times}
\usepackage{latexsym}
\usepackage[T1]{fontenc}
\usepackage[utf8]{inputenc}
\usepackage{microtype}
\usepackage{inconsolata}
\usepackage{graphicx}
\usepackage{amsmath}
\usepackage{amssymb}
\usepackage{booktabs}
\usepackage{multirow}
\usepackage{float}
\usepackage{caption}
\usepackage{placeins}
\usepackage{longtable}

\title{Psychological Constructs in Shared Semantic Space}

\author{
  Hubert Plisiecki \\
  IDEAS Research Institute \\
  \texttt{hplisiecki@gmail.com}
}

\begin{document}
\maketitle

\begin{abstract}
Psychological constructs are often measured in separate instruments, datasets, and research traditions, 
which makes direct comparison difficult. This paper proposes a framework for making such constructs 
semantically commensurate by representing and comparing them as directions in a shared word-embedding space. 
Using Supervised Semantic Differential, we estimate construct-specific semantic gradients from 
text–outcome associations and project them onto theoretically motivated reference axes. As an 
initial test case, we use Valence, Arousal, and Dominance (VAD) as an affective coordinate system. 
First, we recover interpretable VAD directions from English word-level affective norms. Second, 
we project semantic gradients for 27 GoEmotions categories into this space and recover the expected 
organization of emotions, especially along valence and arousal. Third, we apply the same procedure 
to Big Five personality domains and facets derived from IPIP-NEO-300 item–factor associations. 
Domain-level placements are broadly coherent, while facet-level results are more exploratory 
because they rely on sparse questionnaire text. The results suggest that embedding spaces can 
support construct-level comparison across otherwise incommensurable psychological measurements, 
provided that semantic placements are assessed for stability and interpretability.
\end{abstract}

\section{Introduction}
\label{sec:intro}

\footnote{Code repository: \url{https://github.com/hplisiecki/construct\_space}}

Psychological science studies many phenomena that are intuitively related but
empirically difficult to compare. Emotions, personality traits, attitudes,
symptoms, values, and affective states are usually measured with separate
instruments, developed in different research traditions, and scored on
incommensurable scales. As a result, constructs that may be conceptually
adjacent are often analyzed only within the boundaries of a single questionnaire,
dataset, or theoretical framework. This limits our ability to ask broader
questions about how psychological phenomena relate to one another: for example,
whether a personality facet is semantically closer to anxiety, dominance,
curiosity, or positive affect; or how constructs from different studies occupy
a common space of human meaning.

Distributional semantics offers a way to address this problem. Word embeddings
provide a shared representational space in which words, texts, and semantic
directions can be compared geometrically
\citep{mikolov_efficient_2013, kozlowski_geometry_2019}. Rather than treating
embeddings only as predictive features, we use them here as a common semantic
substrate for psychological measurement. The central idea is that constructs
from different studies can be represented as directions in the same embedding
space, making them jointly interpretable even when their original measurements
come from different scales, corpora, or research traditions.

We operationalize this idea using the Supervised Semantic Differential (SSD), a recent method 
inspired by psycholinguistic work on connotative meaning \citep{osgood_measurement_1957} 
that estimates supervised directions in embedding space from paired text representations 
and continuous outcome values \citep{plisiecki_measuring_2025}. Given document embeddings 
and an outcome variable, SSD fits a supervised linear model, back-projects the fitted 
coefficient vector into the original embedding space, and interprets the resulting 
direction through its nearest-neighbor structure. In the present paper, we extend SSD 
from single-construct interpretation to cross-construct comparison by estimating 
multiple construct-specific directions in the same embedding space and projecting 
them onto shared reference axes.

As a first demonstration, we use affective meaning as the reference frame.
Valence, Arousal, and Dominance (VAD) are among the best-validated dimensions of
psychological meaning, with strong grounding in human ratings, affect theory,
and distributional semantics
\citep{warriner_norms_2013, russell_circumplex_1980, russell_evidence_1977}.
We therefore estimate VAD directions in a fixed GloVe embedding space and use
them as interpretable axes onto which other construct gradients can be
projected. Importantly, the contribution is not limited to affect: VAD serves
as a theoretically well-understood test case for the broader claim that
embedding spaces can support shared semantic representations of psychological
phenomena.

We demonstrate the approach across three studies. Study~1 estimates the three
affective reference directions from Warriner et al.'s English word-level VAD
norms \citep{warriner_norms_2013}, yielding unit vectors
$\hat{\boldsymbol{\beta}}_V$, $\hat{\boldsymbol{\beta}}_A$, and
$\hat{\boldsymbol{\beta}}_D$ in GloVe space. Study~2 validates these reference
directions by mapping 27 discrete emotion categories from the GoEmotions corpus
\citep{demszky_goemotions_2020} into the resulting VAD space and comparing the
recovered organization with predictions from the circumplex model of affect
\citep{russell_circumplex_1980}. Study~3 applies the same framework to
personality measurement by deriving semantic gradients from item-level
responses on the IPIP-NEO-300 Big Five inventory
\citep{goldberg_broad-bandwidth_1999} and projecting all five domains and 30 facets
into the affective reference space.
\section{Related Work}
\label{sec:related}

\subsection{Distributional Semantics as Shared Representational Geometry}

Word embeddings represent lexical meaning as points in a continuous vector
space, making semantic relations accessible through distances, neighborhoods,
and directions \citep{mikolov_efficient_2013, pennington_glove_2014}. Although
developed primarily as representations for NLP systems, these spaces have also
been used as analytic objects. Prior work has used embedding geometry to study
semantic change \citep{hamilton_diachronic_2016}, historical shifts in gender
and ethnic stereotypes \citep{garg_word_2018}, and the relational structure of
cultural concepts such as social class \citep{kozlowski_geometry_2019}. Related
projection-based work has shown that linear directions in embedding space can
recover human judgments of object features such as size, danger, and wetness
\citep{grand_semantic_2022}.

These results suggest that embedding spaces can encode interpretable dimensions
of social and psychological meaning. However, most existing work studies a
single lexical contrast, cultural dimension, or historical trajectory. The unit
of analysis is usually a word, dictionary, or predefined contrast. In this
paper, the unit of analysis is a psychological construct. We ask whether
constructs measured in different datasets can be represented as directions in a
common embedding space and meaningfully compared within that shared geometry.

This use case places different requirements on representations than standard
prediction or retrieval. For construct-level measurement, the relevant
properties are not only task accuracy, but whether the space supports stable
linear directions, projection onto interpretable axes, and qualitative audit via
nearest neighbors. This corresponds to the broader distinction between
prediction-oriented and measurement-oriented meaning representations
\citep{plisiecki_prediction-measurement_2026}. Static embeddings are useful here
because they provide a fixed lexical coordinate system in which directions,
projections, and local neighborhoods are directly inspectable. Contextual models
may encode richer information, but their layer dependence and entanglement of
semantic, syntactic, and surface-form signals make them less straightforward as
substrates for the present linear measurement workflow.

\subsection{Affective Meaning and the Semantic Differential}

The affective reference axes used in this paper are motivated by work on
connotative meaning and dimensional theories of affect. The Semantic
Differential represents word meaning through ratings on bipolar scales and
showed that much of this variation can be summarized by a small set of
dimensions, classically evaluation, potency, and activity
\citep{osgood_measurement_1957}. These dimensions map onto how affect theories organize emotion and
connotative meaning in terms of valence, arousal, and dominance
\citep{russell_evidence_1977, russell_circumplex_1980}.
Valence captures the pleasant--unpleasant dimension of meaning; arousal captures
activation level, ranging from calm to excited; and dominance captures perceived
power or control, ranging from submissive to dominant.
Large-scale affective norms provide human ratings of these dimensions for thousands of English lemmas
\citep{warriner_norms_2013}.

We use Valence, Arousal, and Dominance (VAD) as reference axes because they are
well validated, low-dimensional, and theoretically interpretable. The claim is
not that VAD exhausts psychological meaning. Rather, VAD provides a controlled
test case for the broader framework: if constructs can be represented as
directions in a shared embedding space, then they can be projected onto any
theoretically motivated reference axes, as long as it has a significant imprint in the 
embedding space. The space is therefore used as
one interpretable coordinate system for comparing construct gradients, not as
the only possible grounding space.

\subsection{Psychological Constructs and Cross-Instrument Comparability}

Psychological constructs are typically operationalized within specific
instruments, scoring rules, and theoretical traditions. Personality research,
for example, represents individual differences through hierarchical trait
models, including broad Five-Factor Model domains and narrower facets
\citep{mccrae_introduction_1992, john_paradigm_2008}. Public-domain IPIP
instruments provide item-level measures of these domains and facets
\citep{goldberg_broad-bandwidth_1999, goldberg_international_2006}. Such
instruments support within-scale measurement, but they do not provide a common
representation in which constructs from different instruments or datasets can be
directly compared.

Recent work has used text representations to address related problems of
construct comparability, including taxonomic incommensurability in psychological
measurement \citep{wulff_semantic_2025}. Our approach differs in focus. Rather
than comparing questionnaires through aggregate item similarity alone, we
estimate construct-specific semantic gradients from item--outcome associations
and place those gradients in a shared embedding space. This makes the semantic
representation depend not only on item wording, but also on the empirical
relationship between item text and the measured construct.

\subsection{Supervised Semantic Differential}

SSD estimates a semantic gradient from texts paired with a continuous outcome variable 
and interprets the resulting direction
through its nearest-neighbor structure \citep{plisiecki_measuring_2025}. In its
standard form, SSD is used to characterize how meaning varies with one outcome
inside one dataset. The present paper extends this logic to cross-construct
comparison. We estimate multiple construct gradients in the same embedding
space, project them onto shared reference axes, and compare their resulting
coordinates. This turns SSD from a single-construct interpretive method into a
framework for representing and analyzing psychological constructs semantically.

\section{Method: Supervised Semantic Differential}
\label{sec:method}

SSD assumes a collection of documents $d_i$ paired with continuous outcomes
$y_i$. Each document is mapped to a dense vector $\mathbf{x}_i \in \mathbb{R}^D$
via SIF-weighted word embeddings with removal of the top principal component to
reduce anisotropy \citep{mu_all-but--top_2017}. The vectors
are compressed with PCA to $\tilde{\mathbf{x}}_i \in \mathbb{R}^K$ and a linear
model is estimated:
\[
y_i = \alpha + \boldsymbol{\beta}^\top \tilde{\mathbf{x}}_i + \epsilon_i.
\]
The coefficient vector is normalized to unit length to obtain the semantic
gradient $\hat{\boldsymbol{\beta}}$, back-projected to $\mathbb{R}^D$.
The number of PCA components $K$ is selected by a joint interpretability--stability
sweep (AUCK; \citealp{plisiecki_measuring_2025}).

All analyses use GloVe 42B Common Crawl 300-dimensional embeddings
\citep{pennington_glove_2014}, L2-normalized with one component of anisotropy
removed \citep{mu_all-but--top_2017}. Text preprocessing uses spaCy
\texttt{en\_core\_web\_lg} \citep{ines_montani_explosionspacy_2023} with
$a = 10^{-3}$ for SIF weighting.

\section{Study 1: Affective Gradients}
\label{sec:study1}

\subsection{Data}
We use the \citet{warriner_norms_2013} affective norms providing Valence,
Arousal, and Dominance ratings for 13,915 English words on 1--9 scales. Each
word is treated as a single-token document; with \texttt{use\_full\_doc=True}
the document vector equals the word embedding directly. Words not present in
GloVe are excluded. The AUCK sweep searches $K \in \{2, 4, \ldots, 120\}$;
Study~2 uses the same range.

\subsection{Results}

Table~\ref{tab:vad-regression} reports regression statistics at the
sweep-selected $K$ for each dimension. All three axes yield highly significant
fits ($p < 10^{-10}$), confirming that the three affective dimensions have clear,
recoverable geometric structure in GloVe space. Valence achieves the strongest
fit ($r = .73$), Dominance is intermediate ($r = .67$), and Arousal is the
weakest ($r = .58$).

\begin{center}
\small
\begin{tabular}{lrrrrrr}
\toprule
Dimension & $K$ & $R^2_{\text{adj}}$ & $F$ & $p$ & $r$ \\
\midrule
Valence   & 18 & .535 & 876.2 & $<\!10^{-10}$ & .73 \\
Arousal   & 80 & .335 &  87.4 & $<\!10^{-10}$ & .58 \\
Dominance & 54 & .451 & 209.5 & $<\!10^{-10}$ & .67 \\
\bottomrule
\end{tabular}
\captionof{table}{SSD regression results for VAD axis calibration (Warriner et al.\ norms).}
\label{tab:vad-regression}
\end{center}

Appendix~\ref{app:vad-clusters} (Table~\ref{tab:vad-clusters}) reports the full cluster structure at each 
pole of the three calibrated axes. The valence axis separates two positive clusters---aesthetic excellence 
(\emph{stunning, wonderful, exquisite}) and celebration/inspiration (\emph{inspired, celebrate, creative})---from 
two negative clusters of crime/threat (\emph{accusations, criminal, violent}) and moral condemnation 
(\emph{disgusting, heinous, vile}). The arousal axis yields a more fragmented positive side, with clusters 
ranging from frenzied intensity (\emph{rage, frenzy, screaming}) through violent horror (\emph{horrific, terrifying, brutal}) 
and criminal violence (\emph{murder, assault, kidnapping}), in contrast to two compact negative 
clusters of domestic objects (\emph{shelf, drawer, container}) and rural stillness (\emph{cottage, pastoral, meadow}). 
The dominance positive pole combines superlative quality (\emph{wonderful, fantastic, exceptional}) 
with excellence and commitment vocabulary (\emph{dedication, commitment, professionalism}); 
its negative pole encompasses debilitating harm (\emph{severe, crippling, exacerbated}), 
disease and epidemic (\emph{epidemic, plague, cholera}), horrific suffering 
(\emph{horrid, dreadful, ghastly}), and mental crisis (\emph{psychotic, paranoia, suicidal}).
Together the high explained variance, along with the highly face valid gradient clusters prove
that the three affective dimensions are well represented by their corresponding gradients.

\section{Study 2: Mapping Discrete Emotions}
\label{sec:study2}

\subsection{Data and Method}
The GoEmotions dataset \citep{demszky_goemotions_2020} provides 58,009 Reddit
comments, each annotated by multiple raters for 27 emotion categories. For each
emotion, the mean rater-agreement score across annotators ($y_i \in [0, 1]$) is
used as the SSD outcome, with one model fitted per emotion independently using
the same PCA sweep settings as Study~1. Each emotion thus yields a unit gradient
vector $\hat{\boldsymbol{\beta}}_e$ in the same GloVe space as the VAD reference
vectors.

Before projection, each $\hat{\boldsymbol{\beta}}_e$ is orthogonalized against the
normalized mean emotion vector $\hat{\bar{\boldsymbol{\beta}}}$, removing a shared
``generic emotionality'' direction, and renormalized to $\boldsymbol{\beta}_e^{\perp}$.
Affective coordinates are then:
\[
  (v_e,\, a_e,\, d_e) =
  \bigl(\boldsymbol{\beta}_e^{\perp} \cdot \hat{\boldsymbol{\beta}}_V,\;
         \boldsymbol{\beta}_e^{\perp} \cdot \hat{\boldsymbol{\beta}}_A,\;
         \boldsymbol{\beta}_e^{\perp} \cdot \hat{\boldsymbol{\beta}}_D\bigr).
\]

\subsection{Results}

\begin{figure*}[!t]
\centering
\includegraphics[width=\linewidth]{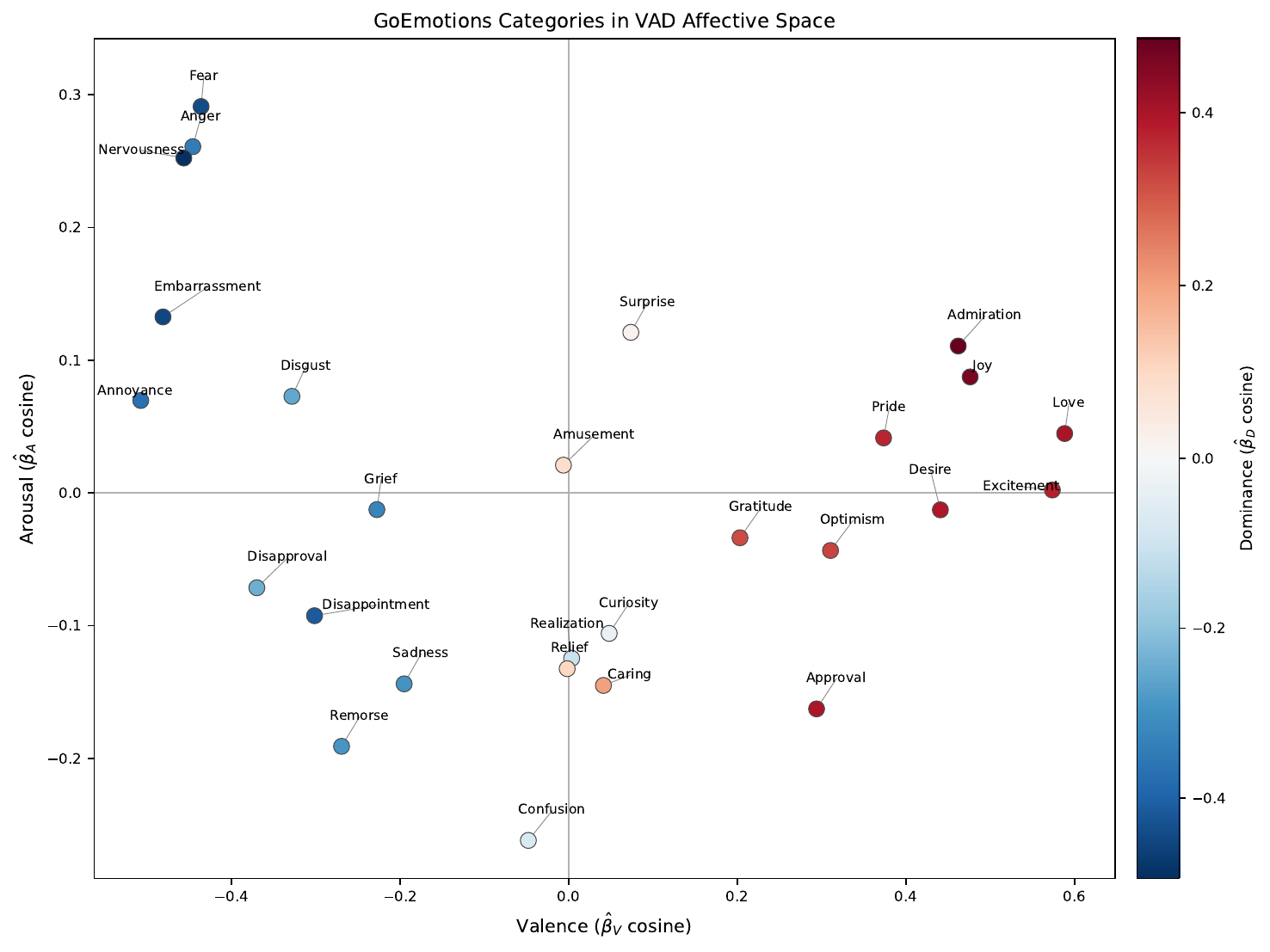}
\caption{VAD coordinates of all 27 GoEmotions categories. Axes are cosine similarities with the calibrated $\hat{\boldsymbol{\beta}}_V$ (x) and $\hat{\boldsymbol{\beta}}_A$ (y) vectors; color indicates $\hat{\boldsymbol{\beta}}_D$ cosine (red = high dominance, blue = low dominance).}
\label{fig:emotion-vad}
\end{figure*}

Table~\ref{tab:ge-regression} reports regression statistics for all 27
emotions. All models reach significance ($p < 10^{-10}$),
though effect sizes vary considerably. Gratitude and Admiration achieve the
highest correlations ($r = .45$ and $.44$ respectively), while low-frequency
or semantically diffuse emotions such as Pride ($r = .13$), Relief ($r = .12$),
and Realization ($r = .14$) show the weakest fits.

\begin{center}
\small
\begin{tabular}{lrrrr}
\toprule
Emotion & $K$ & $R^2_{\text{adj}}$ & $F$ & $r$ \\
\midrule
Admiration     &  30 & .193 &  457.9 & .44 \\
Amusement      & 120 & .176 &  103.7 & .42 \\
Anger          &  30 & .124 &  271.3 & .35 \\
Annoyance      &  30 & .097 &  206.8 & .31 \\
Approval       &  24 & .029 &   72.5 & .17 \\
Caring         &  72 & .073 &   63.4 & .27 \\
Confusion      &  30 & .027 &   54.1 & .17 \\
Curiosity      &  40 & .019 &   29.2 & .14 \\
Desire         &  26 & .019 &   44.1 & .14 \\
Disappointment &  48 & .050 &   64.7 & .23 \\
Disapproval    &  62 & .043 &   42.6 & .21 \\
Disgust        & 116 & .127 &   73.4 & .36 \\
Embarrassment  &  18 & .024 &   79.5 & .16 \\
Excitement     &  16 & .048 &  181.8 & .22 \\
Fear           &  44 & .088 &  127.6 & .30 \\
Gratitude      &  38 & .199 &  376.2 & .45 \\
Grief          &  52 & .023 &   27.4 & .16 \\
Joy            &  32 & .082 &  161.5 & .29 \\
Love           &  16 & .047 &  176.3 & .22 \\
Nervousness    &  34 & .038 &   67.7 & .20 \\
Optimism       &  36 & .058 &   98.5 & .24 \\
Pride          &  26 & .016 &   36.8 & .13 \\
Realization    &  28 & .019 &   39.9 & .14 \\
Relief         &  54 & .013 &   15.2 & .12 \\
Remorse        &  28 & .033 &   72.0 & .18 \\
Sadness        & 120 & .129 &   71.7 & .36 \\
Surprise       & 118 & .079 &   42.6 & .28 \\
\bottomrule
\end{tabular}
\captionof{table}{SSD regression results for all 27 GoEmotions categories. All $p < 10^{-10}$.}
\label{tab:ge-regression}
\end{center}

Figure~\ref{fig:emotion-vad} shows the VAD positions of all 27 emotions. The valence ordering follows 
theoretical predictions closely: joy, admiration, love, and excitement anchor the positive end; 
annoyance, nervousness, embarrassment, anger, and fear anchor the negative end. On the arousal 
axis, emotions associated with activation (anger, fear, nervousness) score higher than 
low-arousal states (relief, sadness, confusion). Dominance (color) closely tracks valence 
($r = .96$ across emotions), largely recapitulating the positive--negative structure without adding
significant independent information. A detrended analysis controlling for this overlap 
(Appendix~\ref{app:detrended-dominance}) reveals that once valence is removed, anger, annoyance, 
and disgust carry relatively higher residual dominance than fear and nervousness among the 
high-arousal negative emotions, consistent with the approach--avoidance distinction in 
dimensional emotion theory. The full numerical coordinates are reported in 
Appendix~\ref{app:ge-vad} (Table~\ref{tab:app-ge-vad}). Cluster tables for all emotion 
gradients can be found in the Appendix~\ref{app:ge-clusters} (Table~\ref{tab:ge-clusters}).

\section{Study 3: Mapping Big Five Personality Domains and Facets}
\label{sec:study3}

\begin{figure*}[!t]
\centering
\includegraphics[width=\linewidth]{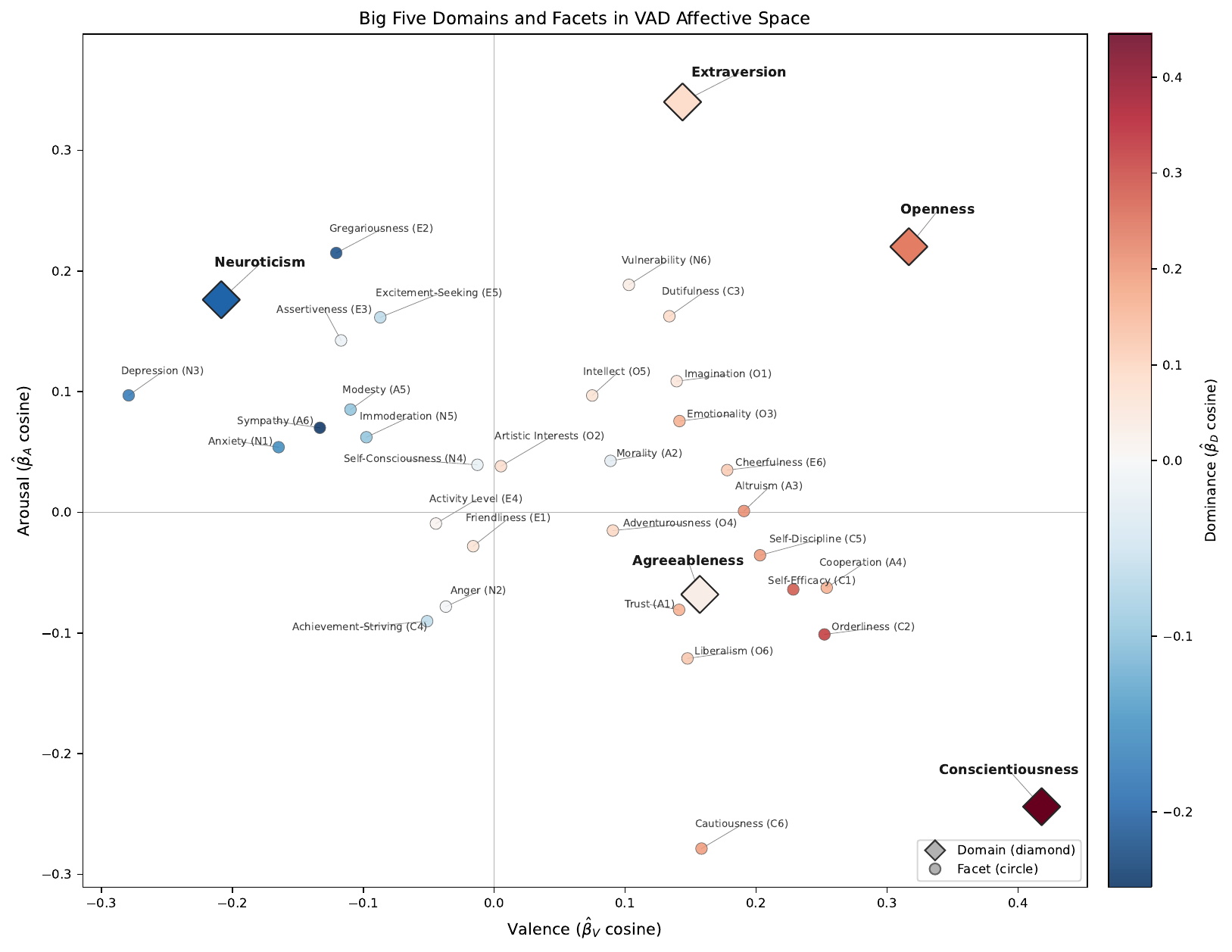}
\caption{Big Five domains (diamonds) and facets (circles) in VAD affective space. Axes are cosine similarities with calibrated $\hat{\boldsymbol{\beta}}_V$ (x) and $\hat{\boldsymbol{\beta}}_A$ (y); color indicates $\hat{\boldsymbol{\beta}}_D$ cosine. Full coordinates in Appendix~\ref{app:ipip-vad}.}
\label{fig:personality-vad}
\end{figure*}

\subsection{Data}
We use the IPIP-NEO-300 dataset \citep{johnson_measuring_2014}, comprising
307,313 participants who completed all 300 items of the NEO Personality
Inventory. Items are scored on a 1--5 Likert scale; approximately half of
items in each domain are negatively keyed and reverse-coded when computing
domain and facet scores. The inventory covers five domains (Neuroticism N,
Extraversion E, Openness O, Agreeableness A, Conscientiousness C) each divided
into six facets of 10 items.

\subsection{Constructing Item-Outcome Variables}
For each domain (60 items) and facet (10 items), a factor score is computed
as the mean of reverse-coded item responses across participants. Pearson $r$
between each item's \emph{original} (pre-reverse-code) responses and the factor
score is computed across all 307,313 participants, yielding the SSD outcome
variable. Using original rather than reverse-coded responses preserves
text--label alignment: negatively-keyed items receive a negative $r$
consistent with their wording, and positively-keyed items retain a positive $r$.
The AUCK sweep searches $K \in \{2, 4, \ldots, K_{\max}\}$ where
$K_{\max} = \min(N_{\text{items}} - 2,\; 30)$ --- capped at 30 rather than 120
because the small item counts (60 for domains, 10 for facets) make higher
$K$ prone to overfitting.

\subsection{Results}

Domain gradients are projected onto the VAD axes via cosine similarity, as in Study~2 (Section~\ref{sec:study2}).
Table~\ref{tab:ipip-domain-regression} reports regression statistics for the
five domains. Four of five reach significance; Agreeableness is non-significant
($p = .27$), likely reflecting the semantic heterogeneity of its items
(prosocial warmth items mixed with items referencing others' suffering).

\begin{center}
\small
\begin{tabular}{lrrrrr}
\toprule
Domain & $K$ & $R^2_{\text{adj}}$ & $F$ & $p$ & $r$ \\
\midrule
Conscientiousness & 20 & .27 & 2.10 & .023 & .72 \\
Openness          & 14 & .27 & 2.55 & .009 & .67 \\
Neuroticism       & 12 & .28 & 2.89 & .005 & .65 \\
Extraversion      &  2 & .18 & 7.21 & .002 & .45 \\
Agreeableness     & 14 & .06 & 1.26 & .269 & .54 \\
\bottomrule
\end{tabular}
\captionof{table}{SSD regression results for the five Big Five domains (sign-consistent method).}
\label{tab:ipip-domain-regression}
\end{center}

Figure~\ref{fig:personality-vad} plots the five domain positions (squares) alongside all 30 facets (circles), 
with dominance encoded as color. The valence ordering C $>$ O $>$ A $>$ E $>$ N is theoretically coherent: 
Conscientiousness reaches the highest valence ($V = +0.42$) and dominance ($D = +0.45$) with the lowest 
arousal ($A = -0.24$), placing it in the calm-control region. Neuroticism occupies the 
negative-valence/positive-arousal/negative-dominance region consistent with activated negative affect. 
Extraversion is distinguished by the highest arousal ($A = +0.34$).

Facets generally cluster within their parent domain's region of the space, providing qualitative 
support for the hierarchical structure of the Big Five. However, facet-level placements should be 
interpreted with caution: each model is fit on only 10 items, and only a minority of the 30 facet 
regressions reach conventional significance (see Appendix~\ref{app:ipip-facets}, Table~\ref{tab:app-ipip-facets}). 
The facet coordinates reported here and in Appendix~\ref{app:ipip-vad} (Table~\ref{tab:ipip-vad}) are 
exploratory and subject to greater sampling uncertainty than the domain-level estimates. Full 
numerical coordinates are provided in Table~\ref{tab:ipip-vad}.

Appendix~\ref{app:ipip-clusters} (Table~\ref{tab:ipip-clusters}) presents the full 
cluster structure at each pole of the five domain-level semantic gradients. The 
Conscientiousness positive pole splits into quality assurance vocabulary (\emph{ensure, achieve, optimal}) 
and product excellence descriptors (\emph{high-quality, robust, sturdy}); its negative pole captures 
impulsivity and disorder (stupidity/disrespect, behavioral incidents). The Neuroticism positive pole 
encompasses online hostility/chaos (\emph{evil, doom, chaos, madness}) and explicit emotional distress 
(\emph{anger, fear, sadness, rage}), while its negative pole groups budgeting and comfort/luxury vocabulary. 
The Openness positive pole is notably diverse---spanning hedonic pleasure, erotic fantasy, and culinary 
appreciation---while both negative clusters index political and legislative vocabulary, reflecting the 
Liberalism facet's co-occurrence with political discourse in the GloVe training corpus. The Extraversion 
positive pole groups playful and joyous vocabulary (\emph{fun, playful, lighthearted}) alongside enthusiasm 
and passion terms; its negative pole is dominated by procedural adverbs 
(\emph{efficiently, smoothly, properly, comfortably}), another corpus artifact rather than a clear 
introversion signal. Agreeableness is omitted here as its domain-level model did not reach significance ($p = .27$).

The alignment between these domain gradients and the GoEmotions emotion gradients
from Study~2 is reported in Appendix~\ref{app:emotion-personality}
(Table~\ref{tab:emotion-personality}), providing a direct illustration of the
framework's capacity for cross-construct comparison between otherwise separate
measurement traditions.

\section{Discussion}
\label{sec:discussion}

The three studies support the central premise that word embeddings can provide
a shared representational space for psychological constructs. Rather than
treating constructs only as scale scores within isolated instruments, SSD
represents them as semantic gradients in a common geometry. This makes it
possible to compare emotion categories, personality domains, and personality
facets using the same operations: projection onto reference axes,
neighborhood inspection, and cross-construct positioning; and regardless of whether they
were measured during the same study. The contribution is
therefore not only a mapping of constructs into affective space, but a broader
framework for making constructs from different studies semantically comparable.

The VAD analyses provide an initial validation of this framework. Study~1 shows
that Valence, Arousal, and Dominance can be recovered as interpretable semantic
directions in GloVe space, although their strength varies. Study~2 then shows
that emotion gradients estimated from GoEmotions occupy theoretically meaningful
positions when projected onto these axes: positive emotions cluster on the
positive-valence side, whereas fear, anger, nervousness, embarrassment, and
annoyance occupy the negative-valence and higher-arousal region. Dominance is
also strongly entangled with valence across emotion categories, which is
theoretically expected: affective dominance, potency, and perceived control 
are often intertwined with evaluative meaning \citep{russell_evidence_1977}. 
The valence detrended residual pattern (Appendix~\ref{app:detrended-dominance})
is modest but interpretable, with anger, annoyance, and disgust showing relatively 
higher dominance than fear and nervousness. This suggests that the dominance gradient is not reducible to
valence alone, even though the two dimensions are strongly coupled.

The personality analysis demonstrates the cross-construct use case more
directly. The Big Five domains come from a different measurement tradition than
emotion labels, yet their SSD gradients can be projected into the same affective
reference space. The resulting domain placements are theoretically coherent:
Conscientiousness occupies a high-valence, high-dominance, low-arousal region,
consistent with its association with important life outcomes
\citep{roberts_power_2007} and with structural accounts emphasizing
industriousness, order, self-control, and responsibility
\citep{roberts_structure_2005}. Neuroticism occupies a negative-valence,
high-arousal, low-dominance region, consistent with work linking Neuroticism to
threat sensitivity and affective reactivity
\citep{robinson_neuroticisms_2025}. Extraversion is distinguished primarily by
high arousal. Furthermore, aligning these domain gradients with the GoEmotions
emotion gradients (Appendix~\ref{app:emotion-personality}) produces interpretable
cross-construct correspondences: Neuroticism with remorse and anger, Extraversion
with joy and love, and Conscientiousness against embarrassment and nervousness.
The facet-level results
are more exploratory, since each facet is estimated from only ten items and only
a minority of facet regressions reach conventional significance. Nevertheless,
they illustrate how shared semantic spaces can expose within-domain
heterogeneity that broad trait scores may obscure.

The central interpretive constraint is that the method estimates the semantics of
construct measurement language, not the latent construct in isolation. In the
personality study, gradients are derived from item texts weighted by their
empirical association with domain or facet scores. The resulting coordinates
therefore describe how a construct is linguistically operationalized in the
instrument. This distinction is important because questionnaire items often
refer to situations, social comparisons, or external negative content that are
not identical to the evaluative meaning of the trait itself. At the same time, 
many psychological constructs are not independent of meaning:
they are partly constituted through shared appraisals, self-descriptions,
behavioral interpretations, and culturally available concepts
\citep{danziger_naming_1997, sparti_making_2001, markus_culture_1991}. For such constructs,
semantic gradients may capture aspects of their latent instantiation that are
not visible from scale scores alone, because they recover the meaning bundle
through which the construct is expressed and understood. More generally, the
artifact-like clusters observed for some personality domains expose an open
methodological problem: questionnaire-derived gradients are estimated from a
small number of short item texts, so limited textual variance can leave the
direction underconstrained and sensitive to accidental lexical or
corpus-specific regularities.

Future work should examine how to stabilize questionnaire-derived construct
gradients. Possible directions include item bootstrapping, stronger
regularization, paraphrase augmentation, larger external item pools, and
alternative representation models. Sentence-level transformer embeddings are a
natural candidate, especially given recent work using such representations to
compare psychological questionnaires \citep{wulff_semantic_2025}. In the
present framework, these embeddings could be regressed onto item--factor
correlations rather than used only for aggregate item similarity, potentially
capturing compositional item meaning more directly. However, contextual
representations introduce their own difficulty: their manifolds may entangle
semantic, syntactic, pragmatic, and surface-form information in ways that make
distilled gradients harder to interpret or compare across constructs. Thus, a
sentence-embedding gradient may appear more stable while making it less clear
whether the signal reflects construct-relevant semantics or model-specific
artifacts. Determining when contextual embeddings improve construct-space
measurement, and when they obscure it, remains an important open question.

\section{Conclusion}
\label{sec:conclusion}

This paper introduced a framework for representing psychological constructs as
semantic gradients in a shared embedding space. Across three studies, we showed
that Valence--Arousal--Dominance axes can be recovered from word-level norms,
that emotion categories from GoEmotions occupy theoretically meaningful
positions when projected onto these axes, and that Big Five personality domains
and facets can be situated in the same affective reference space. More broadly,
the results suggest that embedding spaces can serve as infrastructure for
construct-level comparison: psychological phenomena measured in different
datasets, instruments, and research traditions can be expressed in a common
semantic geometry, enabling comparisons that are not available from isolated
scale scores alone. VAD provides one compact reference frame for this purpose,
but the same approach could be extended to other theoretically motivated
dimensions such as agency, morality, sociality, concreteness, epistemic
certainty, or ideology. In line with measurement-oriented views of
representation quality \citep{plisiecki_prediction-measurement_2026},
construct-space methods should be evaluated not only by whether a gradient fits
its source variable, but also by whether the resulting placement is stable,
interpretable, and supported by nearest-neighbor evidence.

\section*{Limitations}
\label{sec:limitations}

The approach inherits the known limitations of GloVe embeddings, including
insensitivity to negation, context dependence, and biases encoded in the
training corpus. The VAD axes are calibrated on word norms from a convenience
sample primarily of native English speakers; cross-linguistic generalization
is not evaluated here. VAD also provides only one low-dimensional reference
frame for psychological meaning, and its axes should not be treated as
exhaustive or fully independent. In particular, the strong overlap between
Valence and Dominance in the emotion analysis shows that projected coordinates
can reflect shared semantic structure rather than separable psychological
dimensions.

For the Big Five analysis, item-factor correlations are computed within the
IPIP-NEO-300 sample, which over-represents educated, Western, and
English-speaking populations \citep{johnson_measuring_2014}. Facet-level
models are fit with only 10 items, constraining statistical power; confidence
intervals via item-level bootstrapping are a natural next step. More generally,
the method estimates the semantics of construct measurement language, not the
latent construct in isolation. Items referencing negative external content
(e.g., Sympathy) may therefore yield depressed valence estimates that reflect
item phrasing rather than trait valence, a limit of purely lexical semantic
representations.

\section*{Ethical Considerations}

SSD is not designed as a predictive or profiling technology. The gradients
estimated here characterize aggregate semantic regularities in large datasets
and should not be used to make inferences about individuals. More broadly,
semantic placements should not be reified as objective definitions of
psychological constructs or as evidence about the traits, capacities, or values
of particular persons or groups. Because embedding spaces can encode social and
cultural biases from their training corpora, construct gradients may reproduce
or amplify those biases if used without qualitative inspection and external
validation.

All datasets used are publicly available for research purposes. The
IPIP-NEO-300 data are anonymous and collected under informed consent. The text
of this manuscript was partially polished with the assistance of a Large
Language Model; all revised passages were reviewed and corrected by the author.

\bibliography{references}

\appendix

\begin{table}[!t]
\section{VAD Axis Cluster Tables}
\label{app:vad-clusters}
\small
\begin{tabular}{p{0.09\linewidth}clp{0.48\linewidth}}
\toprule
Axis & Pole & $N$ & Cluster Summary (Top Words) \\
\midrule
\multirow{4}{*}{Valence}
 & $+$ & 60 & \emph{Aesthetic excellence}: \textit{stunning, wonderful, superb, exquisite, fabulous, elegant, delightful} \\
 & $+$ & 40 & \emph{Celebration \& inspiration}: \textit{inspired, celebrate, inspiration, enjoy, creative, showcase} \\
\cmidrule{2-4}
 & $-$ & 56 & \emph{Crime \& threat}: \textit{accusations, harassment, crimes, attacks, criminal, violent, abuses} \\
 & $-$ & 44 & \emph{Moral condemnation}: \textit{disgusting, reprehensible, despicable, heinous, shameful, vile} \\
\midrule
\multirow{6}{*}{Arousal}
 & $+$ & 17 & \emph{Frenzy \& rage}: \textit{insane, screaming, rage, frenzy, hardcore, extreme, raging} \\
 & $+$ &  9 & \emph{Unleashed havoc}: \textit{unleash, wreaking, havoc, inflict, unbridled, onslaught} \\
 & $+$ & 22 & \emph{Violent horror}: \textit{horrific, terrifying, brutal, violent, frightening, vicious, deadly} \\
 & $+$ & 16 & \emph{Criminal violence}: \textit{murder, killing, assault, rape, kidnapping, attack, terror} \\
\cmidrule{2-4}
 & $-$ & 59 & \emph{Domestic objects}: \textit{rectangular, tray, flat, circular, shelf, container, drawer, tapered} \\
 & $-$ & 41 & \emph{Rural stillness}: \textit{cottage, pastoral, meadow, cemetery, pasture, composed, adjoining} \\
\midrule
\multirow{6}{*}{Dominance}
 & $+$ & 62 & \emph{Superlative quality}: \textit{wonderful, fantastic, exceptional, excellent, superb, fabulous, unique} \\
 & $+$ & 38 & \emph{Excellence \& commitment}: \textit{creativity, dedication, commitment, professionalism, excellence, generosity} \\
\cmidrule{2-4}
 & $-$ & 38 & \emph{Debilitating harm}: \textit{caused, causing, severe, worse, debilitating, crippling, exacerbated} \\
 & $-$ & 26 & \emph{Disease \& epidemic}: \textit{epidemic, outbreak, cholera, plague, deaths, malnutrition, poisoning} \\
 & $-$ & 17 & \emph{Horrific suffering}: \textit{horrid, horrific, dreadful, horrendous, hideous, ghastly, hellish} \\
 & $-$ & 11 & \emph{Mental crisis}: \textit{psychotic, paranoia, psychosis, hallucinations, suicidal, paranoid} \\
\bottomrule
\end{tabular}
\caption{Cluster structure at the poles of the three calibrated VAD semantic gradients (top-100 neighbors, $k$-means $k\!\in\![2,8]$). $N$ = cluster size. Clusters ordered by centroid--gradient alignment.}
\label{tab:vad-clusters}
\end{table}

\begin{table}[!t]
\section{GoEmotions VAD Coordinates}
\label{app:ge-vad}
\centering\small
\begin{tabular}{lrrr}
\toprule
Emotion & $V$ & $A$ & $D$ \\
\midrule
Love           & $+.59$ & $+.04$ & $+.40$ \\
Excitement     & $+.57$ & $+.00$ & $+.38$ \\
Joy            & $+.48$ & $+.09$ & $+.46$ \\
Admiration     & $+.46$ & $+.11$ & $+.49$ \\
Desire         & $+.44$ & $-.01$ & $+.39$ \\
Pride          & $+.37$ & $+.04$ & $+.37$ \\
Optimism       & $+.31$ & $-.04$ & $+.33$ \\
Approval       & $+.29$ & $-.16$ & $+.40$ \\
Gratitude      & $+.20$ & $-.03$ & $+.32$ \\
Surprise       & $+.07$ & $+.12$ & $+.02$ \\
Curiosity      & $+.05$ & $-.11$ & $-.03$ \\
Caring         & $+.04$ & $-.15$ & $+.20$ \\
Realization    & $+.00$ & $-.12$ & $-.11$ \\
Relief         & $+.00$ & $-.13$ & $+.10$ \\
Amusement      & $-.01$ & $+.02$ & $+.09$ \\
Confusion      & $-.05$ & $-.26$ & $-.08$ \\
Sadness        & $-.20$ & $-.14$ & $-.30$ \\
Grief          & $-.23$ & $-.01$ & $-.33$ \\
Remorse        & $-.27$ & $-.19$ & $-.29$ \\
Disappointment & $-.30$ & $-.09$ & $-.42$ \\
Disgust        & $-.33$ & $+.07$ & $-.26$ \\
Disapproval    & $-.37$ & $-.07$ & $-.24$ \\
Fear           & $-.44$ & $+.29$ & $-.44$ \\
Anger          & $-.45$ & $+.26$ & $-.35$ \\
Nervousness    & $-.46$ & $+.25$ & $-.49$ \\
Embarrassment  & $-.48$ & $+.13$ & $-.45$ \\
Annoyance      & $-.51$ & $+.07$ & $-.37$ \\
\bottomrule
\end{tabular}
\caption{VAD coordinates of all 27 GoEmotions emotion categories (cosine similarity with calibrated axes), sorted by valence.}
\label{tab:app-ge-vad}
\end{table}

\FloatBarrier
\section{Detrended Dominance Analysis}
\label{app:detrended-dominance}

\begin{figure*}[!t]
\centering
\includegraphics[width=\linewidth]{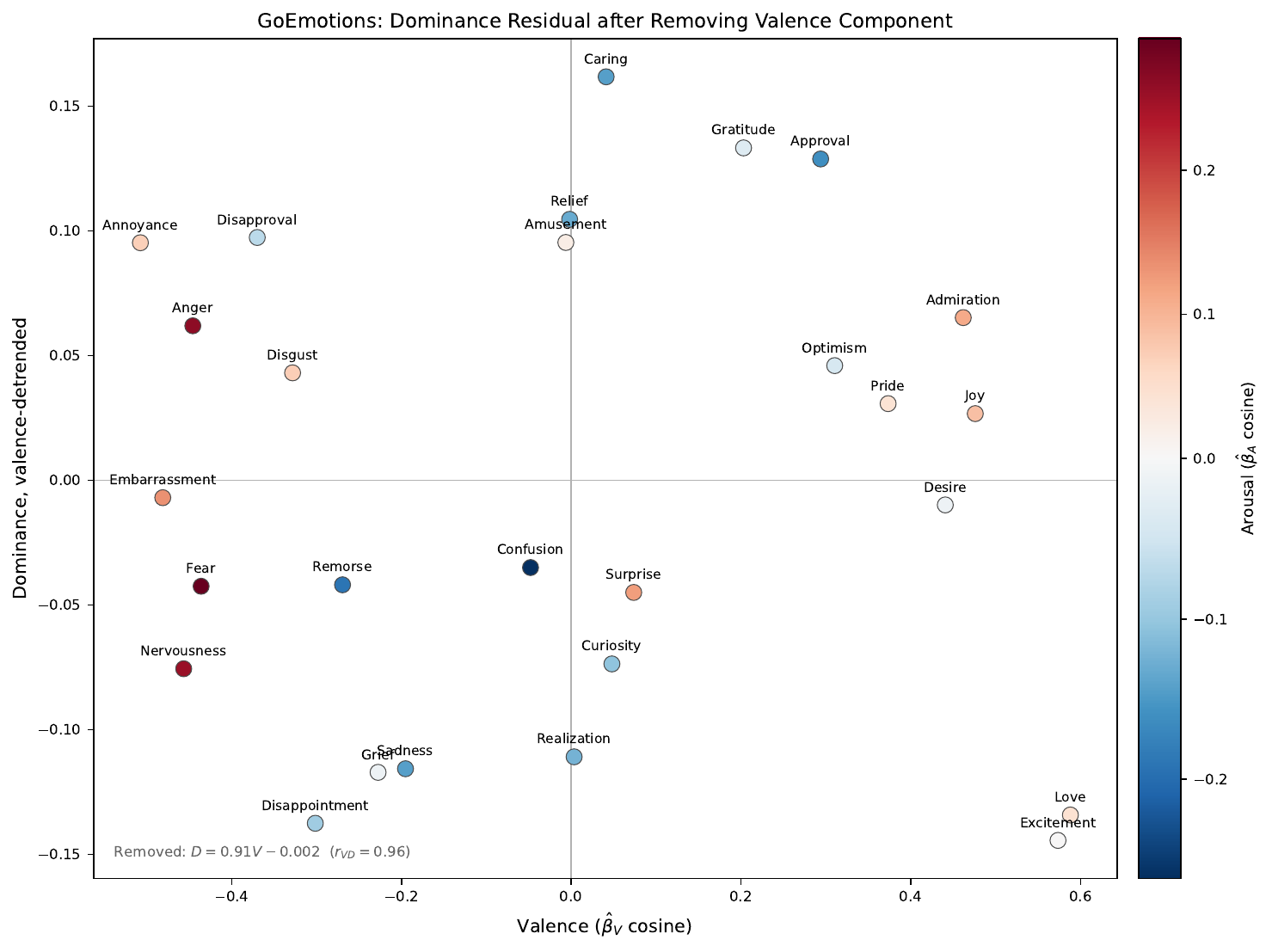}
\caption{Valence versus valence-detrended dominance ($D'$) for all 27 GoEmotions
  categories. Color encodes arousal (red = high, blue = low).
  Among negative high-arousal emotions, anger, annoyance, and disgust carry
  positive $D'$ (relatively dominant given their valence) whereas fear and
  nervousness carry negative $D'$ (relatively submissive), consistent with
  the approach--avoidance distinction.}
\label{fig:emotion-detrended}
\end{figure*}

Raw dominance correlates strongly with valence across the 27 emotion categories
($r = .96$), largely recapitulating the positive--negative structure.
To isolate dominance-specific information, we regress dominance on valence
($D = 0.91 V - 0.002$) and compute residuals $D' = D - \hat{D}$.
Figure~\ref{fig:emotion-detrended} plots valence against $D'$, with arousal
encoded as color.

The key separation is within the negative high-arousal cluster: anger ($D' = +.06$), annoyance ($D' = +.10$), and
disgust ($D' = +.04$) lie above zero, while fear ($D' = -.04$) and nervousness
($D' = -.08$) fall below, consistent with anger being associated with
approach and control and fear with avoidance and submission.
Table~\ref{tab:app-ge-vad-detrended} reports the full detrended coordinates.

\begin{table}[H]
\centering\small
\begin{tabular}{lrrrr}
\toprule
Emotion & $V$ & $A$ & $D$ & $D'$ \\
\midrule
Love           & $+.59$ & $+.04$ & $+.40$ & $-.13$ \\
Excitement     & $+.57$ & $+.00$ & $+.38$ & $-.14$ \\
Joy            & $+.48$ & $+.09$ & $+.46$ & $+.03$ \\
Admiration     & $+.46$ & $+.11$ & $+.49$ & $+.07$ \\
Desire         & $+.44$ & $-.01$ & $+.39$ & $-.01$ \\
Pride          & $+.37$ & $+.04$ & $+.37$ & $+.03$ \\
Optimism       & $+.31$ & $-.04$ & $+.33$ & $+.05$ \\
Approval       & $+.29$ & $-.16$ & $+.40$ & $+.13$ \\
Gratitude      & $+.20$ & $-.03$ & $+.32$ & $+.13$ \\
Surprise       & $+.07$ & $+.12$ & $+.02$ & $-.04$ \\
Curiosity      & $+.05$ & $-.11$ & $-.03$ & $-.07$ \\
Caring         & $+.04$ & $-.15$ & $+.20$ & $+.16$ \\
Realization    & $+.00$ & $-.12$ & $-.11$ & $-.11$ \\
Relief         & $+.00$ & $-.13$ & $+.10$ & $+.10$ \\
Amusement      & $-.01$ & $+.02$ & $+.09$ & $+.09$ \\
Confusion      & $-.05$ & $-.26$ & $-.08$ & $-.04$ \\
Sadness        & $-.20$ & $-.14$ & $-.30$ & $-.12$ \\
Grief          & $-.23$ & $-.01$ & $-.33$ & $-.12$ \\
Remorse        & $-.27$ & $-.19$ & $-.29$ & $-.04$ \\
Disappointment & $-.30$ & $-.09$ & $-.42$ & $-.14$ \\
Disgust        & $-.33$ & $+.07$ & $-.26$ & $+.04$ \\
Disapproval    & $-.37$ & $-.07$ & $-.24$ & $+.10$ \\
Fear           & $-.44$ & $+.29$ & $-.44$ & $-.04$ \\
Anger          & $-.45$ & $+.26$ & $-.35$ & $+.06$ \\
Nervousness    & $-.46$ & $+.25$ & $-.49$ & $-.08$ \\
Embarrassment  & $-.48$ & $+.13$ & $-.45$ & $-.01$ \\
Annoyance      & $-.51$ & $+.07$ & $-.37$ & $+.10$ \\
\bottomrule
\end{tabular}
\caption{VAD coordinates with valence-detrended dominance $D' = D - (0.91 V - 0.002)$, sorted by valence.}
\label{tab:app-ge-vad-detrended}
\end{table}

\clearpage
\onecolumn

\section{GoEmotions Cluster Tables}
\label{app:ge-clusters}

\small
\begin{longtable}{p{0.13\textwidth}clp{0.68\textwidth}}
\caption{Cluster structure at the poles of all 27 GoEmotions semantic gradients.
  $N$ = cluster size; excerpts are the highest-cosine Reddit comments for that cluster.}
\label{tab:ge-clusters}\\
\toprule
Emotion & Pole & $N$ & Cluster Summary (Top Words / Excerpt) \\
\midrule
\endfirsthead
\multicolumn{4}{l}{\small\emph{Table~\ref{tab:ge-clusters} continued from previous page}}\\[2pt]
\toprule
Emotion & Pole & $N$ & Cluster Summary (Top Words / Excerpt) \\
\midrule
\endhead
\midrule
\multicolumn{4}{r}{\small\emph{Continued on next page\ldots}}\\
\endfoot
\bottomrule
\endlastfoot
\multirow{4}{*}{Admiration}
 & $+$ & 74 & \emph{Superlative praise}: \textit{fantastic, amazing, incredible, wonderful, superb, stunning, terrific, remarkable} --- ``Truly incredible work ;)'' \\*
 & $+$ & 26 & \emph{Adverbial excellence}: \textit{wonderfully, amazingly, fantastically, exceptionally, incredibly, stunningly, superbly, extraordinarily} --- ``Wonderfully said!'' \\*
\noalign{\penalty10000\vskip3pt}
 & $-$ & 34 & \emph{Refusal and blame}: \textit{refusing, refused, refuse, refuses, blame, dismiss, ignore, withdraw} \\*
 & $-$ & 37 & \emph{Complaints and worry}: \textit{complaining, complain, bother, worried, bothering, worry, whining, whine} \\
\midrule
\multirow{4}{*}{Amusement}
 & $+$ & 55 & \emph{Humour and comedy}: \textit{hilarious, funny, laugh, comical, jokes, humorous, amusing, joke} \\*
 & $+$ & 45 & \emph{Laughter tokens}: \textit{hahah, hahaha, hahahaha, haha, hahahah, lmao, hahahahaha, ahaha} \\*
\noalign{\penalty10000\vskip3pt}
 & $-$ & 3 & \emph{Gloom and misery}: \textit{challengesorry, challenge!go!busuu, aas} \\*
 & $-$ & 25 & \emph{Noise artifact}: \textit{wretched, dismal, burdened, lackluster, abysmal, forsaken, abandoning, anemic} \\
\midrule
\multirow{4}{*}{Anger}
 & $+$ & 54 & \emph{Insults and slurs}: \textit{stupid, moron, dumbass, idiot, morons, dumb, idiots, ignorant} --- ``There are racist morons everywhere.'' \\*
 & $+$ & 46 & \emph{Profanity}: \textit{fucking, fuck, ass, fucked, pussy, slut, cock, cunt} \\*
\noalign{\penalty10000\vskip3pt}
 & $-$ & 11 & \emph{Elegant and stylish}: \textit{flavour, flavors, sweetness, earthy, hint, colours, freshness, pairing} \\*
 & $-$ & 20 & \emph{Subtle flavours}: \textit{elegant, understated, distinctive, complemented, complements, subtle, stylish, elegance} \\
\midrule
\multirow{4}{*}{Annoyance}
 & $+$ & 55 & \emph{Insults and profanity}: \textit{stupid, shit, dumb, crap, fucking, idiot, fuckin, bullshit} \\*
 & $+$ & 45 & \emph{Ignorance and bigotry}: \textit{ignorant, hypocritical, idiotic, moronic, bigoted, arrogant, hateful, disrespectful} \\*
\noalign{\penalty10000\vskip3pt}
 & $-$ & 8 & \emph{Decorative objects}: \textit{amethyst, filigree, bracelet, lapis, gemstones, brooch, moonstone, wedgwood} \\*
 & $-$ & 14 & \emph{Emoticons and thanks}: \textit{:), :o), \textasciicircum\_\textasciicircum, congrats, thankyou, congratulations, thnx, xoxo} \\
\midrule
\multirow{4}{*}{Approval}
 & $+$ & 36 & \emph{Logical reasoning}: \textit{theoretically, plausible, logically, insofar, fundamentally, feasible, reasoning, logical} \\*
 & $+$ & 64 & \emph{Formal argumentation}: \textit{therefore, however, necessarily, assume, consider, whether, regard, nevertheless} \\*
\noalign{\penalty10000\vskip3pt}
 & $-$ & 3 & \emph{Extreme negativity}: \textit{wretched, pitiful, abject} \\*
 & $-$ & 5 & \emph{Physical symptoms}: \textit{coughing, wheezing, sneezing, choking, fainting} \\
\midrule
\multirow{4}{*}{Caring}
 & $+$ & 16 & \emph{Compassionate care}: \textit{caring, caregivers, care, compassionate, nurturing, supportive, carers, hospice} \\*
 & $+$ & 13 & \emph{Calming and soothing}: \textit{calm, calming, reassuring, comforting, relaxing, relax, gentle, reassured} \\*
\noalign{\penalty10000\vskip3pt}
 & $-$ & 52 & \emph{Astonishing and absurd}: \textit{astounding, astonishing, horrifying, startling, mind-boggling, awe-inspiring, ludicrous, absurd} \\*
 & $-$ & 48 & \emph{Automotive showcase}: \textit{showcased, debuted, prototype, sported, roadster, avant-garde, previewed, lamborghini} \\
\midrule
\multirow{4}{*}{Confusion}
 & $+$ & 29 & \emph{Faulty inference}: \textit{presume, infer, deduce, wrongly, logically, implying, mistaken, erroneously} \\*
 & $+$ & 39 & \emph{Epistemic hedging}: \textit{explain, assume, referring, whether, suggest, indicate, understood, presumably} \\*
\noalign{\penalty10000\vskip3pt}
 & $-$ & 11 & \emph{Exclamation artifacts}: \textit{!!!!!!!!!, !!!!!!!!!!!, !!!!!!!!, !!!!!!!, !!!!!!!!!!!!, !!!!!!, !!!!!!!!!!!!!, !!!!!!!!!!!!!!} \\*
 & $-$ & 13 & \emph{Gratitude and blessing}: \textit{gratitude, kindness, blessing, encouragement, blessings, dedication, loving, gracious} \\
\midrule
\multirow{4}{*}{Curiosity}
 & $+$ & 28 & \emph{Wondering and inquiry}: \textit{wondered, intrigued, asked, inquired, puzzled, curious, fascinated, pondered} \\*
 & $+$ & 8 & \emph{Academic research}: \textit{anthropological, ethnographic, archaeological, researches, anthropologist, smithsonian, hypotheses, subspecies} \\*
\noalign{\penalty10000\vskip3pt}
 & $-$ & 14 & \emph{Exclamation artifacts}: \textit{!!!!!!!, !!!!!!, !!!!!!!!!, !!!!!!!!, !!!!!, !!!!!!!!!!!, !!!!!!!!!!!!, !!!!} \\*
 & $-$ & 9 & \emph{Informal intensifiers}: \textit{sooooo, soooo, sooo, soooooo, sooooooo, awsome, unbelievably, wonderfull} \\
\midrule
\multirow{4}{*}{Desire}
 & $+$ & 31 & \emph{Aspirational striving}: \textit{strive, nurture, inspire, embrace, rediscover, reconnect, unite, cherish} \\*
 & $+$ & 69 & \emph{Wanting and wishing}: \textit{want, 'll, wish, able, decide, wanting, need, forget} \\*
\noalign{\penalty10000\vskip3pt}
 & $-$ & 15 & \emph{Unnerving stimuli}: \textit{disconcerting, unnerving, startling, disturbing, muffled, shrill, alarming, startled} \\*
 & $-$ & 85 & \emph{Absurdity and mockery}: \textit{laughable, ludicrous, idiotic, absurd, moronic, insulting, pathetic, hypocritical} \\
\midrule
\multirow{4}{*}{Disappointment}
 & $+$ & 31 & \emph{Severe harm}: \textit{badly, terribly, severely, horribly, hampered, hurt, crippled, painfully} \\*
 & $+$ & 18 & \emph{Blame and suffering}: \textit{blamed, plagued, blame, woes, disastrous, suffered, exacerbated, blaming} \\*
\noalign{\penalty10000\vskip3pt}
 & $-$ & 7 & \emph{Administrative jargon}: \textit{authorizing, authorizes, designate, memorandum, convene, confer, proponent} \\*
 & $-$ & 21 & \emph{Distinctive features}: \textit{unique, combines, incorporates, distinctive, utilizes, blend, blends, uses} \\
\midrule
\multirow{4}{*}{Disapproval}
 & $+$ & 58 & \emph{Misleading and unethical}: \textit{misleading, unfair, untrue, absurd, ludicrous, dishonest, unethical, inaccurate} \\*
 & $+$ & 42 & \emph{Counter-argumentation}: \textit{contrary, argue, necessarily, justify, oppose, reject, disagree, imply} \\*
\noalign{\penalty10000\vskip3pt}
 & $-$ & 3 & \emph{Reminiscing}: \textit{recollections, reminiscences, reminiscing} \\*
 & $-$ & 14 & \emph{Family relations}: \textit{niece, granddaughter, grandpa, grandson, grandma, nephew, nephews, nieces} \\
\midrule
\multirow{4}{*}{Disgust}
 & $+$ & 49 & \emph{Awful and nasty}: \textit{awful, smelly, horrible, stinky, nasty, ugly, filthy, rotten} \\*
 & $+$ & 51 & \emph{Disgraceful and shameful}: \textit{disgusting, disgraceful, appalling, shameful, reprehensible, heinous, deplorable, horrid} \\*
\noalign{\penalty10000\vskip3pt}
 & $-$ & 17 & \emph{Serene and tranquil}: \textit{serene, tranquil, calm, contemplative, paradise, solace, respite, repose} \\*
 & $-$ & 19 & \emph{Theoretical scepticism}: \textit{proponents, doubted, theorists, hypothesis, theoretically, sceptical, hypotheses, hypothesized} \\
\midrule
\multirow{4}{*}{Embarrassment}
 & $+$ & 17 & \emph{Uncomfortable and awkward}: \textit{embarrassing, uncomfortable, painful, annoying, unpleasant, awkward, irritating, embarrassment} \\*
 & $+$ & 16 & \emph{Ridiculous and pathetic}: \textit{ridiculous, ludicrous, laughable, idiotic, outrageous, pathetic, moronic, downright} \\*
\noalign{\penalty10000\vskip3pt}
 & $-$ & 66 & \emph{Professional expertise}: \textit{expertise, established, knowledge, partnership, collaboration, excellence, engineering, dedicated} \\*
 & $-$ & 34 & \emph{Gaming and fantasy}: \textit{mc, gaia, disciple, feng, alchemy, blacksmith, shui, co-op} \\
\midrule
\multirow{4}{*}{Excitement}
 & $+$ & 45 & \emph{Festive celebration}: \textit{celebration, weekend, festive, celebrate, festivities, thanksgiving, celebrating, summer} \\*
 & $+$ & 55 & \emph{Wonderful and spectacular}: \textit{wonderful, fantastic, delightful, amazing, exciting, fabulous, unforgettable, spectacular} \\*
\noalign{\penalty10000\vskip3pt}
 & $-$ & 10 & \emph{Incorrect and erroneous}: \textit{incorrect, erroneous, inaccurate, faulty, correct, improper, insufficient, defective} \\*
 & $-$ & 9 & \emph{Assertions}: \textit{assertion, assertions, asserting, asserted, assert, refute, accusation, presumption} \\
\midrule
\multirow{4}{*}{Fear}
 & $+$ & 80 & \emph{Terrifying}: \textit{terrifying, frightening, horrifying, horrific, dreadful, horrible, disturbing, terrible} --- ``There's this which is terrifying to think about'' \\*
 & $+$ & 20 & \emph{Feeling terrified}: \textit{terrified, frightened, scared, fearful, panicked, horrified, fear, alarmed} --- ``That's what I'm terrified of.'' \\*
\noalign{\penalty10000\vskip3pt}
 & $-$ & 10 & \emph{Academic honours}: \textit{honors, graduated, honours, scholarship, alumnus, mathematics, laude, majored} \\*
 & $-$ & 10 & \emph{Philosophical tradition}: \textit{philosophy, tradition, ethos, tenets, pluralism, egalitarian, epistemology, rests} \\
\midrule
\multirow{4}{*}{Gratitude}
 & $+$ & 46 & \emph{Thanks and gratitude}: \textit{thank, grateful, thankful, congratulations, gratitude, glad, thanking, sincerely} \\*
 & $+$ & 27 & \emph{Informal thanks}: \textit{thanx, thanks, congrats, :-), pmhi, pmthanks, amthanks, amhi} \\*
\noalign{\penalty10000\vskip3pt}
 & $-$ & 29 & \emph{Assault and harassment}: \textit{raped, assaulted, harassed, handcuffed, molested, beaten, interrogated, humiliated} \\*
 & $-$ & 20 & \emph{Obsession and delusion}: \textit{obsessed, wannabe, vampires, pretends, deranged, inexplicably, loner, imagines} \\
\midrule
\multirow{4}{*}{Grief}
 & $+$ & 35 & \emph{Suffering and illness}: \textit{suffering, suffered, suffer, illness, debilitating, injuries, devastating, afflicted} \\*
 & $+$ & 36 & \emph{Death and tragedy}: \textit{killed, death, died, dying, victims, victim, murdered, survived} \\*
\noalign{\penalty10000\vskip3pt}
 & $-$ & 29 & \emph{Risque slang}: \textit{naughty, slutty, kinky, hottie, raunchy, lesbo, freaky, saucy} \\*
 & $-$ & 17 & \emph{Bureaucratic procedure}: \textit{bylaws, zoning, by-laws, bylaw, rulemaking, clarification, omb, fcc} \\
\midrule
\multirow{4}{*}{Joy}
 & $+$ & 16 & \emph{Cheerful sociability}: \textit{joyful, joyous, cheerful, playful, cheery, celebratory, boisterous, fun-filled} --- ``Wow! What a joyous companion\dots How wonderful!!!'' \\*
 & $+$ & 30 & \emph{Festive celebration}: \textit{christmas, celebration, birthday, thanksgiving, easter, holiday, xmas, festive} --- ``This is amazing. This is how Christmas should be.'' \\*
\noalign{\penalty10000\vskip3pt}
 & $-$ & 58 & \emph{Misleading claims}: \textit{misleading, allegations, falsely, wrongly, inaccurate, erroneously, alleges, allege} \\*
 & $-$ & 42 & \emph{Removal and damage}: \textit{removed, removing, scratched, unsightly, remove, discolored, traces, discoloration} \\
\midrule
\multirow{4}{*}{Love}
 & $+$ & 21 & \emph{Beloved and adored}: \textit{loved, love, beloved, loves, loving, favorite, lover, adore} \\*
 & $+$ & 18 & \emph{Fairy-tale romance}: \textit{fairy, tale, fairytale, princess, enchanted, fairies, wonderland, romance} \\*
\noalign{\penalty10000\vskip3pt}
 & $-$ & 47 & \emph{Unacceptable and unfair}: \textit{unacceptable, unfair, misleading, ineffective, unnecessary, inaccurate, ludicrous, unreasonable} \\*
 & $-$ & 53 & \emph{Violations and penalties}: \textit{violations, penalties, enforcement, violation, officials, imposed, delays, fines} \\
\midrule
\multirow{4}{*}{Nervousness}
 & $+$ & 51 & \emph{Anxious and fearful}: \textit{anxious, frightened, frustrated, fearful, terrified, scared, worried, agitated} \\*
 & $+$ & 49 & \emph{Physical symptoms}: \textit{nausea, dizziness, headaches, tiredness, discomfort, symptoms, headache, vomiting} \\*
\noalign{\penalty10000\vskip3pt}
 & $-$ & 48 & \emph{Religious scripture}: \textit{scripture, scriptures, teachings, biblical, testament, bible, tradition, scriptural} \\*
 & $-$ & 52 & \emph{Honour and awards}: \textit{honor, award, honour, medal, engraved, excellence, honors, heritage} \\
\midrule
\multirow{4}{*}{Optimism}
 & $+$ & 34 & \emph{Hopeful endeavour}: \textit{regain, endeavors, endeavor, revive, hopeful, momentum, prosperous, embark} \\*
 & $+$ & 11 & \emph{Thanks and good wishes}: \textit{congrats, thanks, glad, luck, thanx, bye, thankful, goodluck} \\*
\noalign{\penalty10000\vskip3pt}
 & $-$ & 37 & \emph{Shouting and screaming}: \textit{yelled, shouted, screamed, yelling, shouting, screaming, angrily, loudly} \\*
 & $-$ & 35 & \emph{Hateful language}: \textit{hateful, vulgar, insulting, sexist, obscene, demeaning, derogatory, disgusting} \\
\midrule
\multirow{4}{*}{Pride}
 & $+$ & 27 & \emph{Grateful and thrilled}: \textit{grateful, fortunate, pleased, thrilled, delighted, impressed, proud, immensely} \\*
 & $+$ & 14 & \emph{Respected and honoured}: \textit{respected, honored, esteemed, honoured, renowned, admired, welcomed, fellow} \\*
\noalign{\penalty10000\vskip3pt}
 & $-$ & 30 & \emph{Technical workarounds}: \textit{disable, delete, deleting, config, redirect, manually, disabling, workaround} \\*
 & $-$ & 70 & \emph{Confusion and typos}: \textit{confusion, incorrect, weird, avoid, unintentional, typos, misuse, typo} \\
\midrule
\multirow{4}{*}{Realization}
 & $+$ & 26 & \emph{Speculation and foresight}: \textit{predict, speculate, comprehend, contemplate, anticipate, examine, foresee, theoretically} \\*
 & $+$ & 74 & \emph{Epistemic hedging}: \textit{actually, perhaps, realize, probably, however, might, think, unfortunately} \\*
\noalign{\penalty10000\vskip3pt}
 & $-$ & 7 & \emph{Informal slang}: \textit{hott, hawt, bangin, punchy, loveable, zesty, channelsend} \\*
 & $-$ & 35 & \emph{Thanks and blessing}: \textit{thanx, congrats, awsome, bro, thankyou, congratulations, cheers, bless} \\
\midrule
\multirow{4}{*}{Relief}
 & $+$ & 43 & \emph{Physical tiredness}: \textit{feeling, tired, thankfully, asleep, woke, stay, staying, awake} \\*
 & $+$ & 16 & \emph{Soothing and relaxing}: \textit{soothing, relaxing, relax, calming, soothe, rejuvenate, rejuvenating, relaxation} \\*
\noalign{\penalty10000\vskip3pt}
 & $-$ & 45 & \emph{Insulting language}: \textit{insulting, idiotic, sexist, ludicrous, disrespectful, racist, absurd, demeaning} \\*
 & $-$ & 12 & \emph{Pop artists}: \textit{nicki, minaj, gaga, rappers, soulja, will.i.am, diss, ashanti} \\
\midrule
\multirow{4}{*}{Remorse}
 & $+$ & 20 & \emph{Apologies and regret}: \textit{sorry, apologize, apologies, apologise, regret, forgive, regrets, dear} \\*
 & $+$ & 18 & \emph{Complaining and blame}: \textit{complaining, complain, complained, complains, blaming, blame, ignored, ignoring} \\*
\noalign{\penalty10000\vskip3pt}
 & $-$ & 36 & \emph{Innovative and pioneering}: \textit{innovative, pioneering, cutting-edge, groundbreaking, ground-breaking, visionary, pioneered, entrepreneurial} \\*
 & $-$ & 64 & \emph{World-class showcase}: \textit{world-class, showcasing, showcases, showcase, combines, exhilarating, eclectic, showcased} \\
\midrule
\multirow{4}{*}{Sadness}
 & $+$ & 40 & \emph{Grief and sorrow}: \textit{grief, sadness, sorrow, anguish, despair, loneliness, heartache, disappointment} \\*
 & $+$ & 60 & \emph{Suffering and misery}: \textit{badly, terribly, sad, suffering, hurt, severely, miserable, unhappy} \\*
\noalign{\penalty10000\vskip3pt}
 & $-$ & 6 & \emph{Mathematical notation}: \textit{integers, multiplying, numerals, permutations, longitude, millimeter} \\*
 & $-$ & 21 & \emph{Costumes and attire}: \textit{attire, donning, donned, garb, sported, costumes, decked, costumed} \\
\midrule
\multirow{4}{*}{Surprise}
 & $+$ & 46 & \emph{Astonished and shocked}: \textit{astonished, shocked, amazed, horrified, stunned, surprised, appalled, disgusted} \\*
 & $+$ & 44 & \emph{Astonishing and extraordinary}: \textit{astonishing, incredible, remarkable, astounding, extraordinary, unbelievable, surprising, startling} \\*
\noalign{\penalty10000\vskip3pt}
 & $-$ & 7 & \emph{Foreign name tokens}: \textit{jia, yin, jie, qi, wid, capricorn, leone} \\*
 & $-$ & 9 & \emph{Medical conditions}: \textit{prognosis, cystic, scoliosis, predisposition, musculoskeletal, hcc, diseased, descendants} \\\end{longtable}

\clearpage
\twocolumn

\clearpage
\onecolumn

\section{Big Five Domain Cluster Tables}
\label{app:ipip-clusters}

\small
\begin{longtable}{p{0.14\textwidth}clp{0.68\textwidth}}
\caption{Cluster structure at the poles of the five Big Five domain semantic gradients
  (top-100 neighbors, $k$-means $k\!\in\![2,8]$). $N$ = cluster size; top two clusters per pole shown.}
\label{tab:ipip-clusters}\\
\toprule
Domain & Pole & $N$ & Cluster Summary (Top Words) \\
\midrule
\endfirsthead
\multicolumn{4}{l}{\small\emph{Table~\ref{tab:ipip-clusters} continued from previous page}}\\[2pt]
\toprule
Domain & Pole & $N$ & Cluster Summary (Top Words) \\
\midrule
\endhead
\midrule
\multicolumn{4}{r}{\small\emph{Continued on next page\ldots}}\\
\endfoot
\bottomrule
\multicolumn{4}{l}{\small $^{\dagger}$Embedding-space artifact; see main text.}
\endlastfoot
\multirow{4}{*}{Conscientiousness}
 & $+$ & 57 & \emph{Quality assurance}: \textit{ensure, achieve, ensuring, deliver, quality, optimal, maintain, optimum} \\*
 & $+$ & 43 & \emph{Product excellence}: \textit{high-quality, robust, durable, elegant, excellent, sleek, sturdy} \\*
\noalign{\penalty10000\vskip3pt}
 & $-$ & 27 & \emph{Stupidity \& disrespect}: \textit{stupid, idiotic, irresponsible, obnoxious, disrespectful, crazy} \\*
 & $-$ & 21 & \emph{Behavioral incidents}: \textit{caused, incident, alleged, outburst, inexplicable, misconduct} \\
\midrule
\multirow{4}{*}{Neuroticism}
 & $+$ & 25 & \emph{Online hostility \& chaos}: \textit{aka, evil, ghost, doom, chaos, dead, madness, lyrics} \\*
 & $+$ & 19 & \emph{Emotional distress}: \textit{anger, fear, feelings, sadness, hatred, rage, emotions, jealousy} \\*
\noalign{\penalty10000\vskip3pt}
 & $-$ & 58 & \emph{Budgeting \& planning}: \textit{afford, spend, ordinarily, accustomed, procure, devote, anticipate} \\*
 & $-$ & 42 & \emph{Comfort \& luxury}: \textit{comfortable, accommodations, comforts, sumptuous, upscale, indulge, unwind} \\
\midrule
\multirow{4}{*}{Extraversion}
 & $+$ & 56 & \emph{Enthusiastic energy}: \textit{enthusiasm, excitement, joy, passion, happiness, optimism, laughter} \\*
 & $+$ & 44 & \emph{Playful joviality}: \textit{fun, playful, joyous, comical, lighthearted, joyful, whimsical} \\*
\noalign{\penalty10000\vskip3pt}
 & $-$ & 44 & \emph{Functional instructions}: \textit{properly, able, effectively, allow, determine, maintain, utilize} \\*
 & $-$ & 28 & \emph{Efficient task completion}: \textit{efficiently, quickly, swiftly, smoothly, rapidly, reliably, promptly} \\
\midrule
\multirow{4}{*}{Openness}
 & $+$ & 28 & \emph{Hedonic pleasure}: \textit{enjoy, fun, exciting, enjoying, fantastic, enjoyable, pleasure} \\*
 & $+$ & 19 & \emph{Erotic fantasy}: \textit{erotic, sensual, kinky, fantasies, steamy, lustful, foreplay} \\*
\noalign{\penalty10000\vskip3pt}
 & $-$ & 49 & \emph{Legislative politics}: \textit{senate, legislative, republican, democrats, legislature, elections} \\*
 & $-$ & 47 & \emph{Political discourse}: \textit{political, liberal, conservative, opposition, democracy, politicians} \\
\midrule
\multirow{4}{*}{Agreeableness}
 & $+$ & 57 & \emph{Civic \& charitable}: \textit{non-profit, community, nonprofit, volunteer, charity, charities} \\*
 & $+$ & 43 & \emph{Compassionate connection}: \textit{hope, love, loved, understand, friends, lives, grateful} \\*
\noalign{\penalty10000\vskip3pt}
 & $-$ & 56 & \emph{Geometric objects}$^{\dagger}$: \textit{curved, curving, angled, downwards, curled, thrusting, protruding} \\*
 & $-$ & 44 & \emph{Antagonism \& opposition}: \textit{opponents, intimidate, thwart, attacking, weaken, discredit, evade} \\
\end{longtable}

\clearpage
\twocolumn

\begin{table*}[!t]
\section{IPIP Facet Regression Statistics}
\label{app:ipip-facets}
\centering\small
\begin{tabular}{llrrrrl}
\toprule
Domain & Facet & $K$ & $R^2_{\text{adj}}$ & $F$ & $p$ & $r$ \\
\midrule
\multirow{6}{*}{Conscientiousness}
 & Self-Efficacy       & 2 & .42 & 4.19 & .064 & .74 \\
 & Orderliness         & 2 & .28 & 2.77 & .130 & .66 \\
 & Dutifulness         & 2 & $-.10$ &  .60 & .576 & .38 \\
 & Achievement-Str.    & 2 & .42 & 4.32 & .060 & .74 \\
 & Self-Discipline     & 2 & .01 & 1.03 & .406 & .48 \\
 & Cautiousness        & 2 & .22 & 2.28 & .173 & .63 \\
\midrule
\multirow{6}{*}{Openness}
 & Imagination         & 2 & $-.10$ &  .57 & .587 & .38 \\
 & Artistic Interests  & 2 & .51 & 5.72 & .034 & .79 \\
 & Emotionality        & 2 & .53 & 6.05 & .030 & .80 \\
 & Adventurousness     & 2 & $-.00$ &  .99 & .420 & .47 \\
 & Intellect           & 2 & .43 & 4.39 & .058 & .75 \\
 & Liberalism          & 2 & $-.27$ &  .05 & .953 & .12 \\
\midrule
\multirow{6}{*}{Agreeableness}
 & Trust               & 2 & .51 & 5.77 & .033 & .79 \\
 & Morality            & 2 & .09 & 1.46 & .295 & .54 \\
 & Altruism            & 2 & .11 & 1.57 & .273 & .56 \\
 & Cooperation         & 2 & $-.30$ &  .08 & .929 & .16 \\
 & Modesty             & 2 & .19 & 2.06 & .198 & .61 \\
 & Sympathy            & 2 & .06 & 1.28 & .336 & .52 \\
\midrule
\multirow{6}{*}{Extraversion}
 & Friendliness        & 2 & .41 & 4.07 & .067 & .73 \\
 & Gregariousness      & 2 & .12 & 1.60 & .269 & .56 \\
 & Assertiveness       & 2 & .17 & 1.90 & .219 & .59 \\
 & Activity Level      & 2 & $-.18$ &  .38 & .697 & .34 \\
 & Excitement-Seeking  & 2 & .49 & 5.31 & .040 & .78 \\
 & Cheerfulness        & 2 & .48 & 5.10 & .043 & .77 \\
\midrule
\multirow{6}{*}{Neuroticism}
 & Anxiety             & 2 & .09 & 1.46 & .295 & .54 \\
 & Anger               & 2 & $-.28$ &  .03 & .972 & .09 \\
 & Depression          & 2 & .13 & 1.67 & .255 & .57 \\
 & Self-Consciousness  & 2 & $-.04$ &  .82 & .479 & .44 \\
 & Immoderation        & 2 & .00 & 1.02 & .408 & .48 \\
 & Vulnerability       & 2 & .53 & 6.08 & .030 & .80 \\
\bottomrule
\end{tabular}
\caption{SSD regression statistics for all 30 IPIP-NEO facets (sign-consistent method). All facets use $K{=}2$ (10 items each).}
\label{tab:app-ipip-facets}
\end{table*}

\FloatBarrier
\begin{table}[!t]
\section{Big Five Domain and Facet VAD Coordinates}
\label{app:ipip-vad}
\centering\small
\begin{tabular}{@{}llrrr@{}}
\toprule
Code & Facet & $V$ & $A$ & $D$ \\
\midrule
\multicolumn{2}{@{}l}{\textbf{Conscientiousness}} & $\mathbf{+.42}$ & $\mathbf{-.24}$ & $\mathbf{+.45}$ \\
C1 & Self-Efficacy       & $+.23$ & $-.06$ & $+.28$ \\
C2 & Orderliness         & $+.25$ & $-.10$ & $+.32$ \\
C3 & Dutifulness         & $+.13$ & $+.16$ & $+.09$ \\
C4 & Achievement-Str.    & $-.05$ & $-.09$ & $-.06$ \\
C5 & Self-Discipline     & $+.20$ & $-.04$ & $+.20$ \\
C6 & Cautiousness        & $+.16$ & $-.28$ & $+.19$ \\
\midrule
\multicolumn{2}{@{}l}{\textbf{Openness}} & $\mathbf{+.32}$ & $\mathbf{+.22}$ & $\mathbf{+.23}$ \\
O1 & Imagination         & $+.14$ & $+.11$ & $+.06$ \\
O2 & Artistic Interests  & $+.01$ & $+.04$ & $+.08$ \\
O3 & Emotionality        & $+.14$ & $+.08$ & $+.16$ \\
O4 & Adventurousness     & $+.09$ & $-.02$ & $+.10$ \\
O5 & Intellect           & $+.07$ & $+.10$ & $+.06$ \\
O6 & Liberalism          & $+.15$ & $-.12$ & $+.12$ \\
\midrule
\multicolumn{2}{@{}l}{\textbf{Agreeableness}} & $\mathbf{+.16}$ & $\mathbf{-.07}$ & $\mathbf{+.03}$ \\
A1 & Trust               & $+.14$ & $-.08$ & $+.16$ \\
A2 & Morality            & $+.09$ & $+.04$ & $-.03$ \\
A3 & Altruism            & $+.19$ & $+.00$ & $+.21$ \\
A4 & Cooperation         & $+.25$ & $-.06$ & $+.16$ \\
A5 & Modesty             & $-.11$ & $+.09$ & $-.10$ \\
A6 & Sympathy            & $-.13$ & $+.07$ & $-.24$ \\
\midrule
\multicolumn{2}{@{}l}{\textbf{Extraversion}} & $\mathbf{+.14}$ & $\mathbf{+.34}$ & $\mathbf{+.08}$ \\
E1 & Friendliness        & $-.02$ & $-.03$ & $+.06$ \\
E2 & Gregariousness      & $-.12$ & $+.21$ & $-.22$ \\
E3 & Assertiveness       & $-.12$ & $+.14$ & $-.02$ \\
E4 & Activity Level      & $-.04$ & $-.01$ & $+.02$ \\
E5 & Excitement-Seeking  & $-.09$ & $+.16$ & $-.07$ \\
E6 & Cheerfulness        & $+.18$ & $+.04$ & $+.12$ \\
\midrule
\multicolumn{2}{@{}l}{\textbf{Neuroticism}} & $\mathbf{-.21}$ & $\mathbf{+.18}$ & $\mathbf{-.20}$ \\
N1 & Anxiety             & $-.16$ & $+.05$ & $-.15$ \\
N2 & Anger               & $-.04$ & $-.08$ & $-.00$ \\
N3 & Depression          & $-.28$ & $+.10$ & $-.17$ \\
N4 & Self-Consciousness  & $-.01$ & $+.04$ & $-.02$ \\
N5 & Immoderation        & $-.10$ & $+.06$ & $-.10$ \\
N6 & Vulnerability       & $+.10$ & $+.19$ & $+.03$ \\
\bottomrule
\end{tabular}
\caption{VAD coordinates for all five Big Five domains (bold) and 30 facets. Values are cosine similarities with the calibrated $\hat{\boldsymbol{\beta}}_V$, $\hat{\boldsymbol{\beta}}_A$, $\hat{\boldsymbol{\beta}}_D$ axis vectors. Domain values reproduced from Table~\ref{tab:ipip-domain-regression} for reference.}
\label{tab:ipip-vad}
\end{table}

\pagebreak

\section{Emotion--Personality Alignment}
\label{app:emotion-personality}

Table~\ref{tab:emotion-personality} shows the three most and least aligned GoEmotions
categories for each Big Five domain, based on cosine similarity between the orthogonalized
emotion gradients ($\boldsymbol{\beta}_e^{\perp}$, Study~2) and domain gradients
($\hat{\boldsymbol{\beta}}_d$, Study~3).

Neuroticism aligns with remorse and anger, and negatively with desire and approval.
Extraversion aligns with joy and love ($r \approx +.40$), and negatively with disapproval
and confusion. Openness and Extraversion show considerable overlap, both loading on
excitement and joy; curiosity loads weakly ($r = +.02$). Conscientiousness aligns with
desire, caring, and admiration, and negatively with embarrassment ($r = -.32$) and
nervousness ($r = -.29$). Agreeableness shows uniformly weak alignment (range
$[-.24,\, +.18]$), consistent with its non-significant domain regression
(Table~\ref{tab:ipip-domain-regression}).

\begin{table}[H]
\centering\small
\begin{tabular}{@{}lp{0.30\linewidth}p{0.30\linewidth}@{}}
\toprule
Domain & Most aligned & Least aligned \\
\midrule
Neuroticism      & Remorse ($+.20$)\newline Anger ($+.20$)\newline Annoyance ($+.14$)          & Desire ($-.27$)\newline Approval ($-.23$)\newline Curiosity ($-.18$) \\[6pt]
Extraversion     & Love ($+.40$)\newline Joy ($+.40$)\newline Excitement ($+.34$)               & Disapproval ($-.27$)\newline Approval ($-.26$)\newline Confusion ($-.25$) \\[6pt]
Openness         & Excitement ($+.32$)\newline Joy ($+.27$)\newline Desire ($+.25$)             & Disapproval ($-.27$)\newline Annoyance ($-.21$)\newline Disappointment ($-.20$) \\[6pt]
Agreeableness    & Love ($+.18$)\newline Excitement ($+.14$)\newline Gratitude ($+.14$)         & Embarrassment ($-.24$)\newline Fear ($-.12$)\newline Nervousness ($-.11$) \\[6pt]
Conscientiousness & Desire ($+.27$)\newline Caring ($+.23$)\newline Admiration ($+.23$)         & Embarrassment ($-.32$)\newline Nervousness ($-.29$)\newline Fear ($-.21$) \\
\bottomrule
\end{tabular}
\caption{Top and bottom 3 GoEmotions categories by cosine similarity with each Big Five
  domain gradient. Emotion betas are orthogonalized ($\boldsymbol{\beta}_e^{\perp}$,
  mean emotion direction removed) to isolate emotion-specific content. Domain betas are
  not orthogonalized: with only five domains whose theoretical independence is contested,
  removing a mean direction would be arbitrary. Values are cosine similarities.}
\label{tab:emotion-personality}
\end{table}

\end{document}